# Language Identification of Devanagari Poems

Priyankit Acharya[1], Aditya Ku. Pathak[2], Rakesh Ch. Balabantaray[3], Anil Ku. Singh[4]

[1,2,3]Dept. of CSE IIIT Bhubaneswar, [4]Dept. of CSE IIT BHU

priyankit.97@gmail.com, adityapathak.cse@gmail.com, rakeshbray@gmail.com, aksingh.cse@iitbhu.ac.in

**Abstract**

Language Identification is a very important part of several text processing pipelines. Extensive research has been done in this field. This paper proposes a procedure for automatic language identification of poems for poem analysis task, consisting of 10 Devanagari based languages of India i.e. Angika, Awadhi, Braj, Bhojpuri, Chhattisgarhi, Garhwali, Haryanvi, Hindi, Magahi, and Maithili. We collated a corpora of poems of varying length and studied the similarity of poems among the 10 languages at the lexical level. Finally, various language identification systems based on supervised machine learning and deep learning techniques is applied and evaluated.
.

## 1. Introduction

The task of identifying the language of a provided textual content, is called language identification. Automatic language identification is feasible because of non-randomness of natural languages and they each have uniformity in the usage of characters or character sequences. It is an essential requirement for several Natural Language Processing (NLP) tasks, like, question answering, sentiment analysis, information retrieval and part-of-speech tagging. One such NLP task is poetry computational analysis, although the field remains mostly unexplored but it is getting more and more popular with each passing day. In dealing with poetry, the text classification techniques, however efficient in the case of prose, often have to be remodelled and fine tuned in a very distinct manner. We try to delve into the data and made complete utilization of the lexical content of each poem of the corpus to discover the language to which it belongs. We take poems from 10 different Devanagari script-based languages as a source for our experiments in language identification system for poems. Devanagari based languages are most widely spoken by the people of South Asia as Devanagari being a northern brahmic script is connected to the likes of many South Asian scripts along with Bengali, Gujarati, and Gurmukhi, and, more distantly, to some of South East Asian scripts including Balinese, Thai, and Baybayin. Over 120 spoken Indo-Aryan languages, most notably Hindi, Nepali, Marathi, Maithili, Awadhi, Newari and Bhojpuri uses the Devanagari script.

The goal of our work is to try out existing supervised machine learning and deep larning models to create a forefront language identification system for poem analysis task on Devanagari script-based languages. Prior to building the system, we compute the lexical similarity among all the above-mentioned languages, to get the idea of the dataset. All these methods are discussed and explained broadly in the later sections of the paper.

## 2. Literature Survey

With the advent of several classifiers, language identification was considered to be a solved NLP problem, whose main application is discriminaton between different languages by showcasing almost perfect results, when trained with character and word - level features. But, in the past few years, the state of being unable to reproduce similar results has opened up new questions for the field and kickstarted fresh efforts for finding out the solution for this problem. There have been hardly any work done on language identification of poems but in general some notable works from recent past are as follows :

(Grefenstte, 1999) examined two methods on language identification i.e. trigrams and short words and discovered that trigrams work better for small sentences whereas bigram perform well for longer sentences.

(Murthy & Kumar, 2006) proposed language identification problem as supervised machine learning problem, where the features extracted from a training corpus are used for classification.

(Bhargava & Kondrak, 2010) discussed about a procedure based on SVM in language identification with counts of n-grams as feature sets.

(Vatanen, Vayrynen & Virpioja, 2010) compared two distinct well suited methods for LID task: a Naive Bayes Classifier based on character n-gram models, and the ranking method by Cavnar and Trenkle . The accuracy was found to decrease remarkably when the detected text gets shorter

(Indhuja et al, 2014) discussed on identification of 5 Devanagari based languages : Sanskrit, Hindi, Nepali, Bhojpuri, and Marathi, using n-grams of character and word. The system gave a best performance of 88%.

(Goutte et al, 2016) worked on two types of Portuguese (Brazilian and European) identification system. Journalistic texts were used for their experiments. The system gave an accuracy of 99.5% with character n-grams. Later, a some what identical method was used for classification of Spanish texts by using part-of-speech features, along with character and word n-grams.

(Xu, Wang & Li, 2016) worked on the development for 6 varieties of Mandarin Chinese language identification system. They trained a linear SVM using n-grams of word and character along with word alignment features. The best system gave an accuracy of 82%.

(Kumar et al, 2018 ) tried to develop an automatic language identification system for 5 closely related Indian Languages. Linear SVM was trained using 5-fold cross-validation using character & word n-grams features and also combined n-gram features. The system with feature combination of character bi-gram to 5-grams gave the best result of 96.48%.

## 3. Corpus

There is a lack of accessibility of corpus on Devanagari script-based poems. We constructed our own dataset by scraping poems of 10 Devanagari languages from Kavitakosh[1], a website containing a huge amount of poems across different Indic languages. Overall 1500 poems of famous poets of all times i.e. 150 poems of each language of different poets was accumulated. The dataset was automatically labelled with language information as the scraping was done in terms of languages. The texts were saved in UTF-8 encoding and finally 10 corporas were available for the experiments Finally 10 corporas were available for the experiments

Text pre-processing is a significant and primary step in NLP. In this process inconsistencies in the text are removed. So, we applied data sanitization techniques. We removed punctuation symbols, numerical digits and other Unicode characters beyond Devanagari Unicode range, from the text. Stop-word removal and lemmatization/stemming techniques were not applied as they would alter the phoneme sequences in an n-gram approach, which we have taken as one of the features for our supervised machine learning models.

The general corpus statistics is as follows :

| Number of classes | 10 |
|---|---|
| Number of objects | 1500 |

Table 1: General corpus statistics

The corpus statistics for each language is summarised in Table 2[2] .

## 4. Text Similarity

Text similarity allows us to observe the commonality existing in between the text. No prior study has been done on the similarity of these Devanagari languages from a poetic perspective. So we wanted to investigate it using the data that we compiled. Also since we wanted to build a language identification system, a general study of the lexical similarity of Devanagari these languages would assist us speculate what might be the productive and useful way of tackling the problem.

We have taken Cosine similarity, Euclidean distance and Jaccard index as the text similarity measures.

*Consine Similarity*: It calculates the cosine of the angle between two vectors or in mathematical terms dot product between two vectors. The result of average similarity across the dataset is summarised in Table 3[3].

*Euclidean Distance*: It is the straight-line distance between two points in Euclidean space. Each vector representation could be assumed as a point in a N-dimensional space and the distance between two of such points gives an idea how far or near they are relative to other strings. The result of average euclidean distance across the dataset is summarised in Table 4[4].

*Jaccard Index*: A similar statistic, the Jaccard distance, is a measure of how similar two sets are. Jaccard similarity is for comparing two binary vectors (sets) compared to cosine similarity which is for comparing two real valued vectors. We have treated our bag-of-words vector as a binary vector, where a value 1 indicates a word's presence and 0 indicates a words absence. The result of average jaccard index across the dataset is summarised in Table 5[5].

## 5. Evaluation measures

In this work, F1 score, precision, recall and accuracy are used for the comparison of langauge identification approaches. After the confusion matrix is computed the following terms are found out :

TruePositives (TP) : accurately classified +ve samples
TrueNegatives (TN): accurately classified -ve samples
FalseNegatives (FN):accurately classified +ve samples
FalsePositives (FP) : accurately classified -ve samples

Based on these terms the following evaluation measure are quantified as follows :

---

[1] http://kavitakosh.org/

[2,3,4,5,6,7,8] Tables are mentioned after the references section

Precision = TP / ( TP + FP )

Recall = TP / ( TP + FN )

Accuracy = ( TP + TN ) / ( TP + TN + FP + FN )

F1 score = 2* Precision*Recall / ( Precision + Recall )

## 6. Language Identification: Experiments and Results

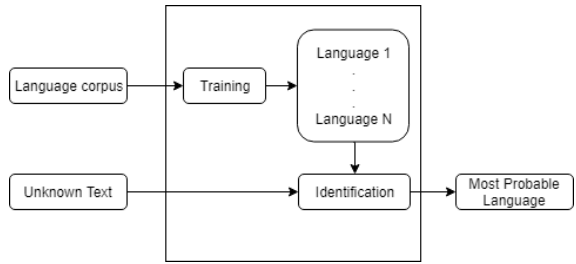

Figure 1: Steps of language identification

In our language identification system, language corpus i.e. the training data is given as input to the model. The model gets trained based on its general rule. And then provided a test data it identifies the most probable language it belongs to.

We have applied and compared supervised machine learning and deep learning models that are suitable for raw textual inputs.

### 6.1 Supervised Machine Learning Techniques

Supervised Machine Learning techniques for Natural Language Processing include a variety of statistical methods that aid in classification and prediction tasks. We have applied the following techniques:

*Support Vector Machine :* The idea behind SVM is that it constructs a line or hyperplane for separating the data into classes. We have SVM with linear kernel for the experiment.

*K-Nearest Neighbors :* KNN is a non-parametric technique. It categorizes a test sample using the label of the majority of examples among its k-nearest i.e. most similar neighbors in the training set based on specific distance metric. We have taken k=3 and euclidean distance as distance metric.

*Random Forest :* Random forests are bagged decision tree models that split on a subset of features on each split.

*Gaussian Naive Bayes :* Naive Bayes is a probabilistic method whose characteristic assumption is to consider that the value of a specific feature is independent of the value of any other feature, given the class variable. When extended to real-valued attributes mostly by gaussian distribution, its Gaussian Naive Bayes.

For developing a sentence level language identification system for 10 Devanagari languages, we divide the dataset into train : test ratio of 75:25. In the experiments we tune various parameters of the classifiers to achieve the best result possible. For training the supervised machine learning models the features used along with the results are as follows:

*Phonemes:* A phoneme is one of the units of sound that distinguish one word from another in a particular language. There have been work done which uses phonemes for NLP tasks (Chourasia et al, 2007). Poets in particular pay attention to the phonemes and how they can be manipulated into musical arrangements (poetic lines). In Devanagari script phonemes are same as that of its letters. The order of letters in Devanagari is based on phonetic principles, which takes both the manner and place of articulation of the vowels and consonants they represent, into consideration.
In Devanagari, vowels can be classified into 2 types: dependent and independent vowels. The dependent form (matra) is used to indicate that a vowel is attached to a consonant. The independent form is used when the vowel occurs alone, at the beginning of a word, or after another vowel.
We used Phonemes (vowels + consonants), Vowels (Independent), Consonants, Matras (Signs) and Phonemes + Matras in our experiment. Out of which the combination of phonemes and matras as a feature set performed better than the rest and the result for all classifiers can be observed in Table 6.

|  | Precision | Recall | F1 | Accuracy |
|---|---|---|---|---|
| **SVM** | 0.6068 | 0.6006 | 0.5975 | 0.6026 |
| **KNN** | 0.5435 | 0.5103 | 0.5104 | 0.5013 |
| **Random Forest** | 0.5650 | 0.5410 | 0.5411 | 0.5306 |
| **Gaussian Naive Bayes** | 0.5423 | 0.3300 | 0.3453 | 0.3306 |

Table 6: Results of classification on (phonemes + matras)

*Tf-Idf:* Tf-Idf vectorization of a corpus of text documents allocates each word in a document a numerical value that is proportional to its frequency in the document and inversely proportional to the number of documents in which it occurs. Thus it's basically a numerical statistic that is intended to reflect how important a word is to a document in a corpus.
We used unigram Tf-Idf, bigram Tf-Idf and trigram Tf-Idf in our experiments. Unigram Tf-Idf performed better than the rest and the result for all classifiers can be observed in Table 7.

|  | Precision | Recall | F1 | Accuracy |
|---|---|---|---|---|
| **SVM** | 0.7926 | 0.7901 | 0.7896 | 0.7893 |
| **KNN** | 0.7893 | 0.7756 | 0.7702 | 0.7706 |
| **Random Forest** | 0.6962 | 0.6976 | 0.6774 | 0.6906 |
| **Gaussian Naive Bayes** | 0.8603 | 0.8438 | 0.8429 | 0.8400 |

Table 7: Results of classification on unigram Tf-Idf

*Character n-gram features:* This is a basic frequency distribution of n characters. We used character bigrams (CB), trigrams (CT), four-grams (C4) and five-grams (C5) and their different combinations in our experiments.

The accuracy top 3 features that performed better for each classifier i.e. SVM, KNN, Random Forest and Gaussian Naive Bayes is described in following Tables:

| **CB+CT+C4** | 0.8461 |
|---|---|
| **C4** | 0.8293 |
| **CU+CB+CT** | 0.825 |

Table 8: Accuracy of top 3 features in SVM

| **CU+CB+CT** | 0.6420 |
|---|---|
| **CB** | 0.6133 |
| **CB+CT** | 0.6106 |

Table 9: Accuracy of top 3 features in KNN

| **C5** | 0.8160 |
|---|---|
| **C4** | 0.7946 |
| **CT** | 0.7706 |

Table 10: Accuracy of top 3 features in Random Forest

| **CB+C5** | 0.8296 |
|---|---|
| **CU+C5** | 0.8293 |
| **C5** | 0.8240 |

Table 11: Accuracy of top 3 features in Gaussian Naïve Bayes

*Word n-gram features:* This is a basic frequency distribution of n words. We used word unigrams (WU), bigrams (WB), trigrams (WT), four-grams (W4) and five-grams (W5) and their different combinations in our experiments.

The accuracy top 3 features that performed better for each classifier i.e. SVM, KNN, Random Forest and Gaussian Naive Bayes is described in following Tables:

| **WU+WB** | 0.6853 |
|---|---|
| **WU+WB+WT** | 0.6453 |
| **WU+WB+WT+W4** | 0.5760 |

Table 12: Accuracy of top 3 features in SVM

| **WU+WB** | 0.1652 |
|---|---|
| **WU+WB+WT** | 0.1506 |
| **WU+WB+WT+W4** | 0.1226 |

Table 13: Accuracy of top 3 features in KNN

| **WU+WB+WT** | 0.5866 |
|---|---|
| **WU+WB** | 0.5706 |
| **WU+WB+WT+W4** | 0.5413 |

Table 14: Accuracy of top 3 features in Random Forest

| **WU+WB+WT** | 0.6906 |
|---|---|
| **WU+WB** | 0.6613 |
| **WU+WB+WT+W4** | 0.6133 |

Table 15: Accuracy of top 3 features in Gaussian Naïve Bayes

*Combined features:* We also experimented with different combinations of both character n-gram features and word n-gram features.

Results of the best combined feature set for each classifier are described in Table 16[6].

As we could see, CT+WT give the best result of 85.21% with SVM classifier. However, the word n-gram features do not seem to perform well as compared to character n-grams. Individually, WU+WB+WT give a below par maximum accuracy of 69.06% with Gaussian naive bayes classifier compared to CB+CT+C4 which gives an accuracy of 84.61% with SVM linear classifier.

The confusion matrix of the best performing classifier i.e SVM for the feature set CT+WT, on the test dataset is summarised in Table 17[7] and language-wise performance for the same is summarised in Table 18:

|  | Precision | Recall | F1-score |
|---|---|---|---|
| **Angika** | 0.97 | 0.89 | 0.93 |
| **Awadhi** | 0.81 | 0.75 | 0.78 |
| **Bhojpuri** | 0.86 | 0.86 | 0.86 |
| **Braj** | 0.74 | 0.78 | 0.76 |
| **Chhattisgarhi** | 0.79 | 0.94 | 0.86 |
| **Garhwali** | 0.87 | 0.92 | 0.89 |
| **Haryanvi** | 1.00 | 0.89 | 0.94 |
| **Hindi** | 0.91 | 0.67 | 0.78 |
| **Magahi** | 0.77 | 0.94 | 0.85 |
| **Maithili** | 0.89 | 0.97 | 0.93 |
| **Avg / Total** | **0.86** | **0.86** | **0.86** |

Table 18: Language wise performance of SVM classifier for (CT+WT) feature.

## 6.2 Deep Learning Techniques

Deep learning and artificial neural networks are manifesting very impressive results at text classification problems, successfully bringing out state of the art outcomes on a suite of standard academic benchmark problems.

For the experiment, we have built a vocabulary index and mapped each word to an integer between 0 and 46,639 i.e. the vocabulary size. Then each sentence becomes a vector of integers.

We have used dropout as method to regularise the neural networks. The idea behind dropout is simple, it randomly disables a fraction of its neurons. This prevent neurons from co-adapting and forces them to learn individually useful features.

During training phase of our neural networks, we iterate over batches of our data of size 256. We have tested the accuracy of our neural networks over 200 and 500 epochs or iterations.

We have used Sequential way of building deep learning networks. Basically, the sequential methodology allows to easily stack layers into the network without worrying too much about all the tensors (and their shapes) flowing through the model. We divide the dataset into train:test ratio of 75:25.

A neural network with more than hidden layer is reffered to as deep neural network. So deep neural used for our experiment are Multi Layer Perceptron (MLP), Long Short Term Memory (LSTM) and Convolutional Neural Network (CNN).

*MLP (Multi Layer Perceptron) :*
MLP is an artificial neural network consists of more than one perceptron. It has an input layer to get the signal, an output layer to make a decision on the input, and in between thses 2 layers, thera are *n* number of hidden layers of the MLP which acts as main computational engine of MLP.
We have used a fully connected network structure with 4 layers. We have used sigmoid activation function on the 2 hidden layers of the network and softmax function in output layer.

*LSTM (Long Short Term Memory) :*
LSTM is type of Recurrent Neural Network (RNN) that manages the contextual information of inputs by learning when to remember and when to forget.
In our LSTM model, the first layer of the network embeds words into low-dimensional vectors. Then there are 2 stacked LSTM layers with added dropout regularization, which classify the result using a softmax layer.

*CNN (Convolutional Neural Network) :* CNN is one of the type of deep neural networks used mostly in the field of Computer Vision. Many studies have shown that CNN is very effective in extracting information from characters or words and encoding it into neural representations (Yin et al, 2017). One of the main reasons for CNNs being unrivalled to traditional machine learning techniques is the ability of CNNs to automatically learn values for their filters based on the task we want the model to perform.
In our CNN model the first layer of the network embeds words into low-dimensional vectors. The next layer performs convolutions over the embedded word vectors using multiple filter sizes. For example, sliding over 3, 4 or 5 words at a time. Next, we globally max-pooled the result of the convolutional layer into a long feature vector, then added dropout regularization, and classified the result using a softmax layer.

The results of deep learning techniques are described in Table 19.

|  | Epochs | Accuracy |
|---|---|---|
| Multi Layer Perceptron | 200 | 0.1747 |
|  | 500 | 0.1973 |
| Long Short Term Memory | 200 | 0.6667 |
|  | 500 | 0.7707 |
| Convolution Neural Networks | 200 | 0.8400 |
|  | 500 | 0.8660 |

Table 19 : Accuracy of deep learning techniwues for epochs : 200 & 500

We could see that, after 500 epochs of training CNN is outperforming rest of the models by giving a accuracy of 86.60%. The confusion matrix of the CNN after 500 epochs is summarised in Table 20[8] and language-wise performance is summarised in Table 21.

|  | Precision | Recall | F1-score |
|---|---|---|---|
| Angika | 0.89 | 0.97 | 0.93 |
| Awadhi | 0.97 | 0.95 | 0.96 |
| Bhojpuri | 1.00 | 0.95 | 0.97 |
| Braj | 0.81 | 0.97 | 0.88 |
| Chhattisgarhi | 0.71 | 0.84 | 0.77 |
| Garhwali | 0.97 | 0.56 | 0.71 |
| Haryanvi | 0.86 | 0.80 | 0.83 |
| Hindi | 0.82 | 0.97 | 0.89 |
| Magahi | 0.76 | 0.88 | 0.82 |
| Maithili | 0.76 | 0.94 | 0.84 |
| **Avg / Total** | **0.86** | **0.86** | **0.86** |

Table 21: Language wise performance of CNN after 500 epochs

## 7. Conclusion and scope for future work

In this work, we have discussed about the creation of Devanagari poems dataset for language identification system for computational poem analysis task of 10 Devanagari script-based languages – Angika, Awadhi, Bhojpuri, Braj, Chhattisgarhi, Garhwali, Haryanvi, Hindi, Magahi and Maithili. Then, based on the corpus we have presented a basic analysis of text similarity using – Cosine Similarity, Euclidean Distance and Jaccard Index to have an idea on text similarity of the accumulated corpus. After that we have experimented with many traditional supervised machine learning algorithms using various features sequences and then with deep neural neyworks. In supervised machine learning the best accuracy obtained was, 0.8521 using a feature combination of CT+WT with linear SVM classifier. In case of deep learning the best accuracy obtained was, 0.8660 with CNN after 500 epochs.

Further we will be taking forward this work by including diphones as a feature for supervised machine learning methods. There is also a scope for implementation of unsupervised machine learning algorithms and various other hybrid deep learning techniques on the corpus, which would provide more insights to the task.

## 8. Acknowledgements

Authors would like to thank NLPR (Natural Language Processing Research) Lab, IIT Bhu, Varanasi for providing with technical assistance to carry out the work mentioned in this paper.

| Sr. No. | Language | Words | Characters(w/o space) | Avg. Word length | Short Words (<=3) | Long Words (>=7) |
|---|---|---|---|---|---|---|
| 1 | Angika | 17,201 | 72,559 | 4.2 | 6,643 | 1,612 |
| 2 | Awadhi | 15,825 | 65,611 | 4.1 | 6,713 | 1,704 |
| 3 | Bhojpuri | 17,742 | 65,149 | 3.7 | 9,334 | 1,066 |
| 4 | Braj | 10,283 | 40,574 | 3.9 | 4,821 | 827 |
| 5 | Chhattisgarhi | 21,051 | 81,160 | 3.9 | 9,837 | 1,596 |
| 6 | Garhwali | 19,136 | 76,028 | 4 | 8,679 | 1,881 |
| 7 | Haryanvi | 28,050 | 1,05,557 | 3.8 | 13,456 | 1,410 |
| 8 | Hindi | 23,003 | 87,334 | 3.8 | 11,707 | 1,827 |
| 9 | Magahi | 19,440 | 73,791 | 3.8 | 9,385 | 1,117 |
| 10 | Maithili | 17,531 | 71,993 | 4.1 | 7,341 | 1,656 |
|  | TOTAL | 1,89,262 | 7,39,756 | 3.93 | 87,916 | 14,696 |

Table 2: Corpus statistics for different languages

|  | Angika | Awadhi | Bhojpuri | Braj | Chhattisgarhi | Garhwali | Haryanvi | Hindi | Magahi | Maithili |
|---|---|---|---|---|---|---|---|---|---|---|
| Angika | 1 | 0.179 | 0.255 | 0.181 | 0.221 | 0.191 | 0.189 | 0.211 | 0.306 | 0.204 |
| Awadhi | 0.179 | 1 | 0.195 | 0.202 | 0.213 | 0.209 | 0.177 | 0.237 | 0.214 | 0.157 |
| Bhojpuri | 0.255 | 0.195 | 1 | 0.352 | 0.407 | 0.355 | 0.362 | 0.414 | 0.481 | 0.335 |
| Braj | 0.181 | 0.202 | 0.352 | 1 | 0.209 | 0.186 | 0.166 | 0.224 | 0.207 | 0.137 |
| Chhattisgarhi | 0.221 | 0.213 | 0.407 | 0.209 | 1 | 0.281 | 0.263 | 0.306 | 0.348 | 0.236 |
| Garhwali | 0.191 | 0.209 | 0.355 | 0.186 | 0.281 | 1 | 0.165 | 0.187 | 0.170 | 0.116 |
| Haryanvi | 0.189 | 0.177 | 0.362 | 0.166 | 0.263 | 0.165 | 1 | 0.271 | 0.235 | 0.172 |
| Hindi | 0.211 | 0.237 | 0.414 | 0.224 | 0.306 | 0.187 | 0.271 | 1 | 0.357 | 0.286 |
| Magahi | 0.306 | 0.214 | 0.481 | 0.207 | 0.348 | 0.170 | 0.235 | 0.357 | 1 | 0.304 |
| Maithili | 0.204 | 0.157 | 0.335 | 0.137 | 0.236 | 0.116 | 0.172 | 0.286 | 0.304 | 1 |

Table 3: Average Cosine Similarity matrix across the dataset

|  | Angika | Awadhi | Bhojpuri | Braj | Chhattisgarhi | Garhwali | Haryanvi | Hindi | Magahi | Maithili |
|---|---|---|---|---|---|---|---|---|---|---|
| Angika | 0 | 11.76 | 14.33 | 10.8 | 21.13 | 12.69 | 15.04 | 22.96 | 40.50 | 10.99 |
| Awadhi | 14.39 | 0 | 13.52 | 10.76 | 18.38 | 17.51 | 15.52 | 19.57 | 26.44 | 7.93 |
| Bhojpuri | 19.15 | 13.52 | 0 | 11.73 | 22.67 | 19.24 | 15.98 | 28.14 | 47.99 | 11.39 |
| Braj | 10.82 | 10.76 | 11.73 | 0 | 19.81 | 16.51 | 16.74 | 22.62 | 30.94 | 7.47 |
| Chhattisgarhi | 21.13 | 18.38 | 22.67 | 19.81 | 0 | 18.61 | 17.08 | 23.84 | 44.46 | 10.43 |
| Garhwali | 12.69 | 17.51 | 19.24 | 16.51 | 18.61 | 0 | 17.76 | 16.11 | 17.15 | 7.79 |
| Haryanvi | 15.04 | 15.52 | 15.98 | 16.74 | 17.08 | 17.76 | 0 | 24.64 | 43.07 | 9.98 |
| Hindi | 22.96 | 19.57 | 28.14 | 22.62 | 23.84 | 16.11 | 24.64 | 0 | 45.57 | 10.93 |
| Magahi | 40.50 | 26.44 | 47.99 | 30.94 | 44.46 | 17.15 | 43.07 | 45.57 | 0 | 12.26 |
| Maithili | 10.99 | 7.93 | 11.39 | 7.47 | 10.43 | 7.79 | 9.98 | 10.93 | 12.26 | 0 |

Table 4: Average Euclidean Distance matrix across the dataset

|  | Angika | Awadhi | Bhojpuri | Braj | Chhattisgarhi | Garhwali | Haryanvi | Hindi | Magahi | Maithili |
|---|---|---|---|---|---|---|---|---|---|---|
| Angika | 1 | 0.0326 | 0.0365 | 0.0358 | 0.0329 | 0.0291 | 0.0268 | 0.0321 | 0.0435 | 0.0312 |
| Awadhi | 0.0326 | 1 | 0.0418 | 0.0546 | 0.0451 | 0.0408 | 0.0359 | 0.0488 | 0.0428 | 0.0326 |
| Bhojpuri | 0.0365 | 0.0418 | 1 | 0.054 | 0.0507 | 0.0473 | 0.0446 | 0.0586 | 0.0555 | 0.0384 |
| Braj | 0.0358 | 0.0451 | 0.054 | 1 | 0.0486 | 0.0450 | 0.0383 | 0.0561 | 0.0462 | 0.0335 |

| | | | | | | | | | |
|---|---|---|---|---|---|---|---|---|---|
| **Chhattisgarhi** | 0.0329 | 0.0451 | 0.0507 | 0.0486 | 1 | 0.0444 | 0.0393 | 0.0491 | 0.0493 | 0.0335 |
| **Garhwali** | 0.0291 | 0.0408 | 0.0473 | 0.0450 | 0.0444 | 1 | 0.0320 | 0.0389 | 0.0348 | 0.0244 |
| **Haryanvi** | 0.0268 | 0.0359 | 0.0446 | 0.0383 | 0.0393 | 0.0320 | 1 | 0.0566 | 0.0467 | 0.0292 |
| **Hindi** | 0.0321 | 0.0488 | 0.0586 | 0.0561 | 0.0491 | 0.0389 | 0.0566 | 1 | 0.0725 | 0.0506 |
| **Magahi** | 0.0435 | 0.0428 | 0.0555 | 0.0462 | 0.0493 | 0.0348 | 0.0467 | 0.0725 | 1 | 0.0367 |
| **Maithili** | 0.0312 | 0.0326 | 0.0384 | 0.0335 | 0.0335 | 0.0244 | 0.0292 | 0.0506 | 0.0367 | 1 |

Table 5: Average Jaccard Index matrix across the dataset

| | Feature | Precision | Recall | F1 | Accuracy |
|---|---|---|---|---|---|
| **SVM** | CT+WT | 0.8612 | 0.8632 | 0.8614 | 0.8521 |
| **KNN** | CB+W5 | 0.6377 | 0.6098 | 0.5982 | 0.6106 |
| **Random Forest** | C4+WU | 0.8299 | 0.8184 | 0.8147 | 0.8133 |
| **Gaussian Naïve Bayes** | C4+WT | 0.8499 | 0.8466 | 0.8289 | 0.8240 |

Table 16: Performance summary of best performing combined feature for each classifier.

| ACTUAL CLASS | PREDICTED CLASS | | | | | | | | | |
|---|---|---|---|---|---|---|---|---|---|---|
| | Angika | Awadhi | Bhojpuri | Braj | Chhattisgarhi | Garhwali | Haryanvi | Hindi | Magahi | Maithili |
| **Angika** | **34** | 0 | 0 | 0 | 0 | 0 | 0 | 0 | 3 | 1 |
| **Awadhi** | 0 | **30** | 0 | 4 | 2 | 1 | 0 | 2 | 1 | 0 |
| **Bhojpuri** | 0 | 0 | **31** | 0 | 1 | 0 | 0 | 1 | 3 | 0 |
| **Braj** | 0 | 3 | 0 | **28** | 2 | 1 | 0 | 0 | 1 | 1 |
| **Chhattisgarhi** | 0 | 1 | 0 | 0 | **33** | 1 | 0 | 0 | 0 | 0 |
| **Garhwali** | 0 | 0 | 1 | 1 | 1 | **34** | 0 | 0 | 0 | 0 |
| **Haryanvi** | 0 | 1 | 0 | 0 | 2 | 0 | **33** | 0 | 0 | 1 |
| **Hindi** | 0 | 1 | 4 | 5 | 1 | 1 | 0 | **31** | 2 | 1 |
| **Magahi** | 1 | 0 | 0 | 0 | 0 | 1 | 0 | 0 | **34** | 0 |
| **Maithili** | 0 | 1 | 0 | 0 | 0 | 0 | 0 | 0 | 0 | **33** |

Table 17: Confusion matrix of SVM classifier for (CT+WT) feature

| ACTUAL CLASS | PREDICTED CLASS | | | | | | | | | |
|---|---|---|---|---|---|---|---|---|---|---|
| | Angika | Awadhi | Bhojpuri | Braj | Chhattisgarhi | Garhwali | Haryanvi | Hindi | Magahi | Maithili |
| **Angika** | **34** | 0 | 0 | 0 | 0 | 0 | 0 | 0 | 1 | 0 |
| **Awadhi** | 0 | **36** | 0 | 0 | 0 | 0 | 2 | 0 | 0 | 0 |
| **Bhojpuri** | 0 | 0 | **37** | 0 | 1 | 0 | 0 | 0 | 1 | 0 |
| **Braj** | 1 | 0 | 0 | **30** | 0 | 0 | 0 | 0 | 0 | 0 |
| **Chhattisgarhi** | 0 | 0 | 0 | 1 | **27** | 0 | 3 | 0 | 1 | 0 |
| **Garhwali** | 2 | 1 | 0 | 5 | 2 | **36** | 0 | 5 | 5 | 8 |
| **Haryanvi** | 0 | 0 | 0 | 0 | 7 | 0 | **32** | 0 | 0 | 1 |
| **Hindi** | 0 | 0 | 0 | 0 | 0 | 0 | 0 | **31** | 1 | 0 |
| **Magahi** | 1 | 0 | 0 | 1 | 0 | 1 | 0 | 1 | **29** | 0 |
| **Maithili** | 0 | 0 | 0 | 0 | 1 | 0 | 0 | 1 | 0 | **29** |

Table 20: Confusion matrix of CNN after 500 epochs